\documentclass{article}

 \usepackage[dblblindworkshop, final, nonatbib]{neurips_2025}
\workshoptitle{Scaling Environments for Agents (SEA)}
\usepackage[numbers]{natbib}

\usepackage[utf8]{inputenc} % allow utf-8 input
\usepackage[T1]{fontenc}    % use 8-bit T1 fonts
\usepackage{hyperref}       % hyperlinks
\usepackage{url}            % simple URL typesetting
\usepackage{booktabs}       % professional-quality tables
\usepackage{amsfonts}       % blackboard math symbols
\usepackage{amsmath}
\usepackage{amssymb}
\usepackage{nicefrac}       % compact symbols for 1/2, etc.
\usepackage{microtype}      % microtypography
\usepackage{xcolor}         % colors
\usepackage{tikz}
\usepackage{tcolorbox}
\usepackage{times}
\usepackage{latexsym}
\usepackage{tikz}
\usepackage{xcolor}
\usepackage[dvipsnames]{xcolor}
\usepackage{booktabs}
\usepackage{geometry}
\usepackage{tabularx}
\usepackage{listings}
\usepackage{caption}
\usepackage{tabularx}
\usepackage{etoolbox}
\usepackage{enumitem}
\usepackage{graphicx}
\usepackage{listings}
\usepackage{pifont}
\usepackage{subcaption}
\definecolor{codebackground}{rgb}{0.95,0.95,0.95}

\definecolor{rootcolor}{RGB}{55, 55, 55}
\definecolor{reasoningcolor}{RGB}{70, 130, 180}
\definecolor{systemcolor}{RGB}{60, 179, 113}
\definecolor{planningcolor}{RGB}{244, 164, 96}
\definecolor{domaincolor}{RGB}{147, 112, 219}

\usepackage[T1]{fontenc}

\usepackage[utf8]{inputenc}

\usepackage{microtype}

\usepackage{inconsolata}
\newcommand{\datasetname}{\textsc{Memtrack}}

\title{\datasetname: Evaluating Long-Term Memory and State Tracking in Multi-Platform Dynamic Agent Environments} % This is a more SEO suitable name. Potentially remove Memtrack from title?

\author{Darshan Deshpande$^{*}$\quad Varun Gangal$^{*}$\quad
Hersh Mehta \quad \textbf{Anand Kannappan} \quad \\ \textbf{Rebecca Qian} \quad \textbf{Peng Wang}\\ 
Patronus AI\\
  {\texttt{\{darshan, varun.gangal, hersh, anand, rebecca, peng\}@patronus.ai}}
}

\begin{document}

\maketitle

\def\thefootnote{*}\footnotetext{These authors contributed equally to this work}\def\thefootnote{\arabic{footnote}}

\begin{abstract}
Recent works on context and memory benchmarking have primarily focused on conversational instances but the need for evaluating memory in dynamic enterprise environments is crucial for its effective application. We introduce {\datasetname}, a benchmark designed to evaluate long-term memory and state tracking in multi-platform agent environments. {\datasetname} models realistic organizational workflows by integrating asynchronous events across multiple communication and productivity platforms such as Slack, Linear and Git. Each benchmark instance provides a chronologically platform-interleaved timeline, with noisy,  conflicting,  cross-referring information as well as potential codebase/file-system comprehension and exploration. Consequently, our benchmark tests memory capabilities such as acquistion, selection and conflict resolution. We curate the {\datasetname} dataset through both manual expert driven design and scalable agent based synthesis, generating ecologically valid scenarios grounded in real world software development processes. We introduce pertinent metrics for Correctness, Efficiency, and Redundancy that capture the effectiveness of memory mechanisms beyond simple QA performance. Experiments across SoTA LLMs and memory backends reveal challenges in utilizing memory across long horizons, handling cross-platform dependencies, and resolving contradictions. Notably, the best performing \textsc{GPT-5} model only achieves a 60\% Correctness score on \datasetname. This work provides an extensible framework for advancing evaluation research for memory-augmented agents, beyond existing focus on conversational setups, and sets the stage for multi-agent, multi-platform memory benchmarking in complex organizational settings. $\datasetname$ instances are available at \footnote{{\datasetname} Dataset: \url{https://drive.google.com/file/d/1ymMXmOIhCUcwC1WKOW8kioZgeYyrt-qe/view?usp=sharing}}
\end{abstract}

\section{Introduction}
Recent advances in  adaptivity of AI agents have surfaced the need for dynamic memory components that can acquire, understand and utilize relevant information from previous interactions~\cite{zhang2024survey}. More specifically, memory has been used to enhance personalization~\cite{li2024hello} and performance of LLM agents in robotics~\cite{glocker2025llm}, financial trading~\cite{li2023tradinggpt},  healthcare~\cite{abbasian2023conversational} and science research~\cite{yan2024efficient}. While these advances bring real-life impact and are scaling to multi-agent systems faster~\cite{wang2025mirix, nan2025nemori}, benchmarking memory in agentic systems has been largely limited to conversational setups. Benchmarks like \texttt{LoCoMo}~\cite{maharana2024evaluating} and \texttt{LongMemEval}~\cite{wu2024longmemeval} are focused on single thread conversational benchmarking but with the adoption of automatic context acquisition techniques~\cite{mei2025survey}, it becomes important to evaluate memory for real-world agentic tasks that require live context switches. 

In ecologically valid~\cite{bowman-dahl-2021-will} tasks, memory can be used for reasoning, managing and evolving enterprise components like codebases and documentation~\cite{joshi2025swe}. In this paper, we structure our evaluation setup as a unique sandboxed environment with realistic situations covering key concepts of contradictory memory management, and multi-hop and cross-knowledgebase content retrieval and reasoning. 
Unlike previous works that depend on specific implementation of memory and inclusion of external components like archival memory~\cite{packer2024memgpt}, \textsc{MemTrack} is agnostic to the backend mechanism for storing and maintaining memory. This makes \textsc{MemTrack} suitable for long context models as well as agentic systems with memory components attached. Furthermore, benchmarks like MemoryAgentBench~\cite{hu2025evaluating} are made on top of existing datasets like $\infty$Bench~\cite{zhang2024infty}, making them non-cohesive and susceptible to biases like the MCQ output format bias~\cite{zheng2023large}.
% Write about answer determinism
% Write about synthetic + manual curation -> scalability

We build a robust set of asynchronous tools including Linear \footnote{\url{https://linear.app} and competitor Jira are well-known ticketing and project tracking platform offerings}, Slack \footnote{\url{https://slack.com}}and live notification servers, along with a dockerized file system and Git (using Gitea\footnote{\url{https://about.gitea.com/}}) to faithfully replicate environments that modern agentic systems live in~\footnote{\url{https://cursor.com/dashboard?tab=integrations}, \url{https://docs.anthropic.com/en/docs/claude-code/mcp}}. {\datasetname} contains 47 carefully curated and novel long context datapoints simulating realistic enterprise SWE workflows. This paper introduces three  complementary data curation methodologies: 1) a top-down manual curation protocol, 2) a scalable, bottom-up, agent-based approach that generates novel memory data points by working backward from closed pull requests in widely used open-source repositories, and 3) a hybrid approach that combines human supervision with automated iterative refinement. To minimize non-deterministic behavior and overfitting on multiple choice QA for end-to-end evaluation, \textsc{MemTrack} supports brief phrase outputs that  can be conveniently evaluated via direct and approximate matching using an LLM-as-judge. 

The research questions we try to answer in this paper are as follows:
\begin{enumerate}
    \item Can modern LLMs reason over large event histories and codebases to answer SWE questions?
    \item Can agents use memory components such as Zep \cite{rasmussen2025zeptemporalknowledgegraph} and Mem0 \cite{chhikara2025mem0} to more effectively reason over and about large event platform timelines?
    \item Can LLM agents optimally access and remember information from across platforms?
\end{enumerate}

Through our findings, we show that state-of-the-art LLMs fail to perform effective multi-platform context reasoning. Furthermore, we show that LLMs cannot use memory tools effectively and using such tools increases redundancy in planning and overall tool use. In our qualitative studies, we find that LLMs often struggle with context retention and need to repeatedly read tickets, conversations and code files throughout the conversation. Thus, {\datasetname} proposes a scalable design for automated data curation for memory benchmarking and surfaces key error patterns in modern LLMs with respect to memory usage.

% \begin{figure*}[!ht]
%     \centering
%     \includegraphics[width=\textwidth]{figures/Enterprensieve_Timelinegen.drawio.png}
%     \vspace{-0.49em}
%     \caption{A sketch of our timeline generation process}
%     \label{fig:timeline_generation}
%     \vspace{-0.8em}
% \end{figure*}

% \begin{figure*}[!ht]
%     \centering
%     \includegraphics[width=\textwidth]{figures/Enterprensieve_Flowchart.drawio.png}
%     \vspace{-0.49em}
%     \caption{The overall flow of how we construct}
%     \label{fig:overall}
%     \vspace{-0.8em}
% \end{figure*}
%Each benchmark instance provides a chronologically interleaved timeline with noisy, conflicting, and cross-referential information, paired with codebase comprehension that require memory capabilities such as acquisition, selection, and conflict resolution. 

\section{Relevant Work}
Memory for agents is an essential component,  especially for applications such as LLM personalization~\cite{shao-etal-2023-character, zhong2024memorybank}, social simulation~\cite{park2023generativeagentsinteractivesimulacra, fink2024ai}, gaming~\cite{wang2023jarvis1openworldmultitaskagents, zhu2023ghostminecraftgenerallycapable, NEURIPS2024_5949a875, hu2025surveylargelanguagemodelbased}. Evaluating agentic systems has primarily focused on four key aspects: Tool usage, Planning, Reflection, and Memory~\cite{yehudai2025survey}. These aspects make agents unique and distinguish them from standard conversational RAG systems~\cite{fan2024survey}.

\begin{figure*}[!ht]
    \centering
    \includegraphics[width=0.99\linewidth]{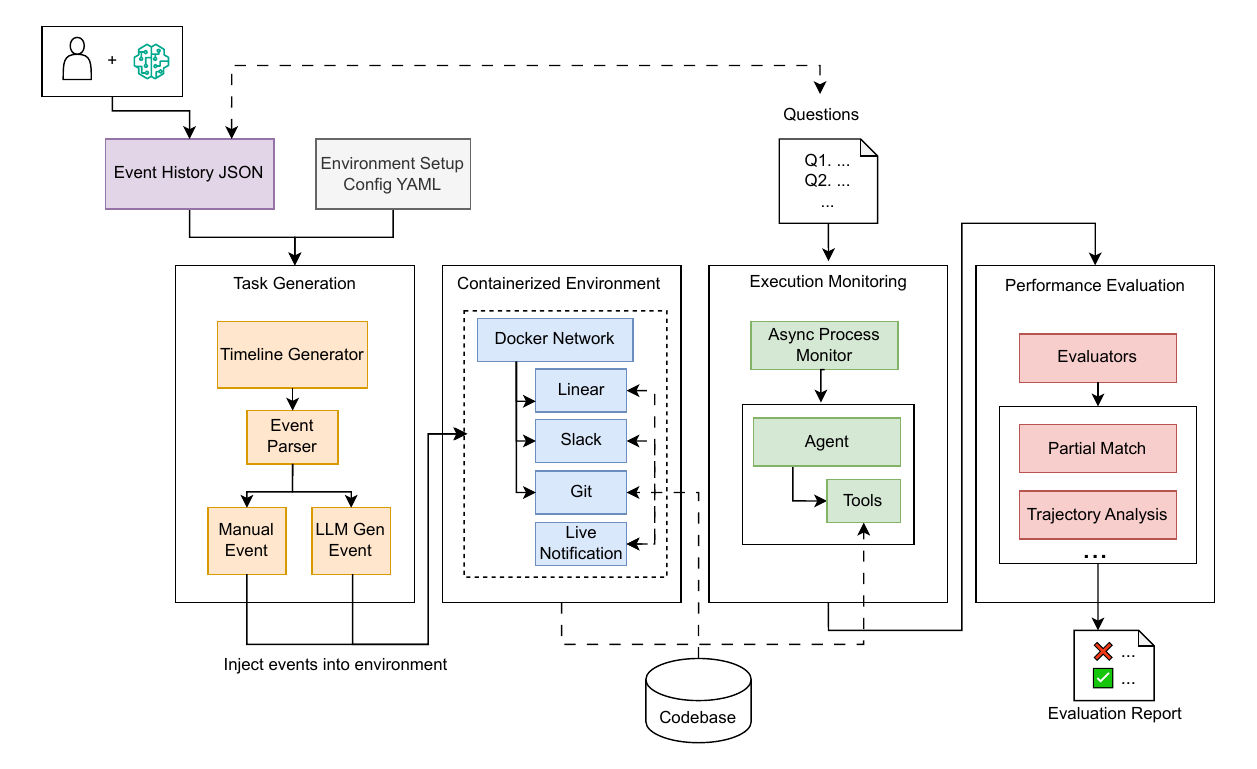}
    \caption{\textsc{MemTrack}'s Data Generation and Evaluation procedure is divided into four parts: task generation, injection of events into containerized environment, agent execution monitoring and sequential question injection, and performance evaluation}
    \label{fig:memtrack}
\end{figure*}

\vspace{-0.8em}
\paragraph{Agentic Evaluation}
% Cover tool use, 
Evaluations for agentic tools revolve around task planning, tool selection, tool invocation, and response generation but existing benchmarks test for capabilities in a fragmented manner~\cite{qu2025tool}. For example, state of the art benchmarks like WTU-Eval~\cite{ning2024wtu}, ToolBench~\cite{qin2023toolllm}, AppWorld~\cite{trivedi2024appworld} and T-Eval~\cite{chen2023t} are restricted in focus to only a subset of these categories. In the sector of evaluating planning capabilities, datasets such as AgentBench~\cite{liu2023agentbench}, $\tau$Bench~\cite{yao2024taubenchbenchmarktoolagentuserinteraction} and WebArena~\cite{zhou2023webarena} have gained popularity but lack ecological validity or are simplistic for real world use-case testing~\cite{huang2024understandingplanningllmagents}. Planning trajectories for agents and their evaluation have been studied for specific tasks such as web search~\cite{lù2025agentrewardbenchevaluatingautomaticevaluations} and while efforts have been made to synthetically generate agentic trajectories~\cite{xu2024agenttrek}, evaluation of techniques for generating long range planning and evaluating such trajectories still remains an open problem. Since memory is often implemented as a wrapper around tools, for example, in the form of a vector database~\cite{packer2024memgpt, chhikara2025mem0} or graphs~\cite{rasmussen2025zeptemporalknowledgegraph}, core concepts of evaluating tools such as tool planning, selection, invocation and response generation can be directly mapped to memory evaluations. One additional component to memory frameworks is the ability of these frameworks to forget or retain information based on requirements~\cite{shan2025cognitivememorylargelanguage} which is understudied. 

\vspace{-0.8em}
\paragraph{Agentic Memory}
% Talk about the categories: acquistion, retrieval, reasoning, maintenance 
\citet{shan2025cognitivememorylargelanguage, du2025rethinkingmemoryaitaxonomy}
show that memory evaluation can be divided into three categories: acquisition, utilization and maintenance. 
To test accuracy of utilization, agentic benchmarks like ~\citet{hu2025evaluating, wu2024longmemeval, maharana2024evaluating} use long-context, needle in the haystack-like tests for retrieval benchmarking. While acquisition of memory is relatively less standardized, works such as Reflexion~\cite{shinn2023reflexionlanguageagentsverbal} and MemInsight~\cite{ salama2025meminsightautonomousmemoryaugmentation} rely on LLM self reflection capabilities to decide which memories are relevant and useful. ~\citet{xu2025amemagenticmemoryllm} utilize concepts of link generation and automatic memory evolution to qualitatively show better t-SNE clusters of memory embeddings. Despite of these efforts, quantitatively evaluating which memories to retrieve has not been thoroughly studied. In parallel, existing works show different methods of memory maintenance. \citet{hu2025evaluating} specifically use existing coarse to fine classification data mappings to simulate test time learning whereas \citet{maharana2024evaluating, li2025memorizationvsreasoningupdating} introduce conflicting information throughout the event history to introduce confusion and promote deep reasoning. \citet{zhong2023memorybankenhancinglargelanguage} cover concepts of long range memory management such as forgetting that are derived from human cognitive concepts.

\paragraph{Agentic Memory Implementations}
\citet{packer2024memgpt} were early adopters of using file systems for storage and retrieval of information across long contexts. In the following years, several other techniques were built that involved databases~\cite{chhikara2025mem0, hu2023chatdb} and graphs~\cite{li2024graphreader, rasmussen2025zeptemporalknowledgegraph} for storing long range information scalably. In parallel, \citet{kang2025memoryosaiagent} revealed that a hierarchical implementation of memory helps performance on the LoCoMo dataset. \citet{zeng2024structural} show that iterative retrieval consistently outperforms other methods across various scenarios and that mixed memory techniques are better than any single memory implementation type. ~\citet{lee2024human} show that simply storing and retrieving gists as memory from long range text can improve performance. Other works~\cite{yoon2024compact, hu2024hiagenthierarchicalworkingmemory} show how hierarchical text compression techniques can help RAG systems reason over long range texts.  from~\citet{wang2024agentworkflowmemory, xu2025amemagenticmemoryllm}  build memory to improve long context performance for specific workflows which suggest that diverse evaluation workflows are needed for better testing of generalizable memory components. This work, \textsc{MemTrack}, is an effort to create complex and diverse agentic workflows by scaling agent search spaces with the help of a variety of tools and tasks.    

\paragraph{Reinforcement Environments for LLM Agents}
The recent shift towards better alignment techniques~\cite{shao2024deepseekmathpushinglimitsmathematical} has resulted in better generalization for different tasks\cite{sun2024easytohardgeneralizationscalablealignment}. Following this progress, the need for unique simulation environments is growing since this becomes an automated source of useful data for improving agent performance~\cite{laleh2024surveyenhancingreinforcementlearning}. Several studies recently have shown that well formed, high quality environments can scale performance beyond simple supervised fine tuning~\cite{zhao2024epohierarchicalllmagents}. One such example of an evaluation environment is AndroidWorld~\cite{rawles2025androidworlddynamicbenchmarkingenvironment} which benchmarks autonomous computer control agents. \textsc{MemTrack} is a step in this direction of creating a benchmark for memory augmented agent systems.

\section{\textsc{MemTrack} Dataset}

The {\datasetname} dataset is developed to answer the important research questions on long-term and cross-platform memory in agentic systems. {\datasetname} dataset includes 47 instances. Each instance includes a series of events, called \textbf{timeline} denoted as $T$, and question-expected answer pairs  $<Q_i, {A_i}^{*}>$. The system is designed such that the evaluated LLM is unaware of the number of question-answer pairs within one instance and these questions are introduced strictly sequentially to remove the possibility of preemptive solution planning. An example of a {\datasetname} instance can be found in~\autoref{tab:instance-example}. 

In {\datasetname}, a timeline represents a series of events in the environment that are accessible by tools or notifications. The event timeline is noted as $T = (E_1, ...E_n)$.  Each event $E_i$ includes a set of attributes, including meta-information such as timestamp, platform types ${e_i}^{P}$ where $P=\{Slack,Linear,Git\}$, and content such as Slack messages. During the creation of the benchmark environments, the timeline is loaded onto the corresponding Linear, Slack and Git servers, and is ready to be accessed by agent memory components. The timeline events will be only accessible through tools set by the servers and are not available to agents as a whole. This necessitates multi-platform context switching, information retention and effective reasoning. At instantiation (as shown in~\autoref{fig:memtrack}), an event parser separates manually curated events from prompts that need to be passed to LLMs to generate an event. If LLM generation is required, we simulate the event data using \textsc{Claude-4-Sonnet} and then continue to inject both event types into the containerized environment. The dataset is compiled through a consolidation of three event history creation approaches.\\\newline
\begin{table}[h!]
\caption{Dataset Metrics: Definitions and Statistics}
\centering
\begin{tabularx}{\textwidth}{p{2.5cm} X c c}
\hline
\textbf{Metric} & \textbf{Description} & \textbf{Mean} & \textbf{Max} \\
\hline
Per-Instance Event Count & Defined as $|E_i|$, the count of events in an instance's timeline $T$. Longer event timelines indicate more turns, longer context, and more challenging to acquire memory and memorize. & 39.9 & 115 \\
\hline
Event Tokens & Measures the total number of tokens in the event timeline $E_i$. Larger tokens are a factor indicative of harder tasks. & 4.01K & 11.1K \\
\hline
Per-Instance Platform Entropy & For each timeline $E_{i}$, we calculate its events' distribution on each platform $\hat{P}(e^{Platform}=p |e \sim E_{i})$, and then compute the entropy $H = -\sum_{p \in P} {\hat{P}}(p|e) \log {\hat{P}}(p|e)$. Higher entropy is desirable because the events are spread more evenly across multiple platforms. & 0.668 & 0.989 \\
\hline
Cross-Platform References & An LLM-judge (rubric in \S\ref{subsubsec:appendix_crossreference}) assigns a soft multi-label 0-1 score to indicate the likelihood of an event $e_i$ referring other platforms $p \in P, p \neq e_i^{Platform}$. An event is called cross-platform-referring, if the score exceeds a threshold (0.3). Higher number of cross-platform-referring events indicates more communication across platforms between the members within an organization, while a lower number indicates the echo-chamber problem \cite{kumar2018community}, which would defeat the goal of evaluating the cross-platform dynamics. & 2.1 & 19 \\
\hline
Chronological Heterophily & Another form of trivialization in cross-platform dynamics can arise through events on each platform occurring in neatly separated sub-phases within a timeline without ever interleaving. To check if this anti-pattern is prevalent, we measure Chronological Heterophily, specifically the probability of an event being followed by an event of another platform. We get the maximum likelihood estimation (MLE) at the per-example level by computing the fraction of times this happens for non-terminal events. We see that we have a healthy number for this at $0.364$. & 0.364 & 0.714 \\
\hline
Timeline Span in Hours & Besides the number of events, another distinct way of gauging the length of a timeline is the timespan it covers in its own frame of reference, which is the difference between the first and last event times $e_{i,|E_{i}|}.\tau - e_{i,1}.\tau$ measured in hours. From \ref{tab:dataset_metrics}, we can see that our timelines span 878 hours on average, stretching across 3000 hours in some cases. & 878 & 3049 \\
\hline
Per-Instance Question Count & Defined as $|Q_i|$, the total number of questions asked in an instance, which includes an initial question and optional follow-up questions. & 3.2 & 5 \\
\hline
\end{tabularx}
\label{tab:dataset_metrics}
\end{table}
\begin{table*}[!ht]
\caption{\small An example of instance, including timeline events and question-answer pairs. Note that the Git events are set up in the filesystem in the Git server and are not shown here. }
    \small
    %\ninesize
    \setlength{\tabcolsep}{2pt}
    \begin{tabular}{l l}
    \toprule %\hline
          \begin{tabular}{l} \textbf{Platform} \end{tabular}  &   \begin{tabular}{l} \textbf{Events} \end{tabular}    \\ \midrule
        
        \begin{tabular}{l} Linear \end{tabular}  & \begin{tabular}{p{0.85\textwidth}}    "timestamp": "20250410T0900", "title": "MLE-Bench Integration for Competition Analysis Pipeline",   "description": "Integrate MLE-Bench framework into our ML competition analysis pipeline to automate competitor analysis and leaderboard tracking. This will enable automated data preparation and performance benchmarking for internal ML competitions.", "team": "ml",  "priority": "high",  "lead": "sarah\_chen", "attached\_resources": [ "/design/mle-bench-integration-spec.pdf", "/research/competition-analysis-requirements.md"]
         \end{tabular}   \\ 
         \midrule %\hline

        \begin{tabular}{l} ... \end{tabular}  & \begin{tabular}{p{0.85\textwidth}}  ...
         \end{tabular}   \\ 
         \midrule %\hline
        \begin{tabular}{l} Slack \end{tabular} & \begin{tabular}{p{0.85\textwidth}} "timestamp": "20250410T1430",  "channel": "\#ml", "sender": "sarah\_chen", "message": "The error is happening during dataset download. Looks like it's in one of the core files - either mlebench/data.py, mlebench/utils.py, or maybe mlebench/registry.py. Import failure in some exception handling code around lines 180-200."
        \end{tabular}   
        \\ \midrule %\hline

        \begin{tabular}{l} Linear \end{tabular} & \begin{tabular}{p{0.85\textwidth}}   "timestamp": "20250410T1630",  "title": "Code Analysis: MLE-Bench Import Error Source Investigation",  "description": "Source code analysis of import failures in MLE-Bench data preparation module. Multiple import statements identified in core data processing files. Issues located in exception handling code sections. Detailed line-by-line analysis required for complete understanding of scope.",  "team": "ml", "priority": "medium", "lead": "sarah\_chen", "status": "done"
        \end{tabular}   
        \\ \midrule 
           
        \begin{tabular}{l} ... \end{tabular} & \begin{tabular}{p{0.85\textwidth}}  ...
        \end{tabular}   
        
            \\ \midrule %\hline
        \begin{tabular}{l} Slack \end{tabular} & \begin{tabular}{p{0.85\textwidth}}   "timestamp": "20250411T1440",  "channel": "\#ml",  "sender": "david\_wong",  "message": "Check the compatibility testing ticket - I documented the specific working version there. The one that passed all our preparation tests."
        \end{tabular}   

        \\ \midrule %\hline
        \begin{tabular}{l} Slack \end{tabular} & \begin{tabular}{p{0.85\textwidth}}  "timestamp": "20250411T1600", "channel": "\#engineering", "sender": "sarah\_chen",  "message": "@alex\_kim Could be related to the new Istio service mesh rollout. I'll check the telemetry data."
        \end{tabular}

        \\ \midrule %\hline
        \begin{tabular}{l} ... \end{tabular} & \begin{tabular}{p{0.85\textwidth}}  ...
        \end{tabular}   
          \\ \bottomrule
       \\
         \begin{tabular}{l} {Questions}  \end{tabular}   
         & \begin{tabular} {p{0.85\textwidth}} \textbf{Question 1}:  Based on sarah\_chen's vague mention of `lines 180-200' and 'exception handling code', clone the repository and examine the source: how many total Python files in the entire mlebench/directory (including subdirectories) contain any import statement that references the kaggle package? \\ \textbf{Question 2}:  Following conversations about 'competition-specific modules' and `experiments directory', investigate the actual repository structure: what is the exact filename (without path) of the Python file in the mlebench/competitions/subdirectory that contains the longest individual function by line count? \\ \textbf{Question 3}: After correlating references to 'data preparation pipeline' and 'dataset download functionality', analyze the repository code: what is the total number of function definitions (def statements) in the mlebench/data.py file? \\ ...
        \end{tabular}  
        \\
        \hline \\
        \begin{tabular}{l} {Answers}  \end{tabular}   
        & \begin{tabular} {l} 
        \textbf{Answer 1}: 9 \\ \textbf{Answer 2}: prepare.py \\ \textbf{Answer 3}: 4 \\ ...
        \end{tabular}  
        \\    
        \bottomrule
    \end{tabular}
    %\vspace{-0.6\abovedisplayskip}
    \label{tab:instance-example}
    %\vspace{-2.5ex}
\end{table*}
\noindent\textbf{Bottom-Up Approach} \\
Ensuring the cohesive utilization of all communication channels and tools in the system, we design a bottom-up approach to creation of data points and unique realistic scenarios. Similar to SWE Bench~\cite{jimenez2024swebench}, we first select a list of popular, open source repositories from GitHub.
% \footnote{\url{https://github.com/search?q=language%3APython+stars%3A%3E500+size%3A%3C10000+is%3Apublic+is%3Aopen+license%3Amit+license%3A0bsd++license%3Aapache-2.0+license%3Agpl+&type=repositories}}
We then design an explorative agent (details available in~\autoref{appendix:bottom-up}) with access to web search tools and bash commands to explore closed issues on each one of these repositories. The agent is tasked to select a unique set of issues that have successfully merged PRs linked to them and preferably discussions leading to the resolution of the issue. The agent is further instructed to focus on smaller scale Pull Requests (PRs)\footnote{Small scale refers to PRs with less than 3 files changed} of up to two changed files that can be deterministically verified. Once the list is curated, two human annotators reviewed the list of extracted issues and resolutions, and agreed on the removal of the issues that do not have clear deterministic answers. After this filtering stage, we create another agent that takes the issues, resolutions and discussions as context and simulates natural slack conversations and Linear tickets pertaining to the issue. Once generated, the agent is instructed to augment the given data with contextualized counterfactual data and irrelevant distractors. We ensure that the distractors remain as grounded as possible to real-world workplace conversational exchanges. The generated timeline is then evaluated manually. Finally,  the task is generated from the created issues and the actual solution implemented in the merged PR. \\\newline
\textbf{Top-Down Approach} \\
We ask a team of 4 in-house experts, who have worked in product and engineering organizations and have experience developing products that require collaborations among multiple teams, to provide examples in their own language, based on their own experience. Those examples represent the scenarios about how the real-world problems are defined, communicated, and eventually solved in an iterative way to address encountered challenges. Then we use their descriptions in the prompt to generate the timeline events using LLMs. Then the experts manually inspect the generated timeline and update it to make the generated timeline align with the task descriptions. The question and answer pairs are manually generated by reviewing generated timelines and ensure that the tasks are related to the task descriptions. \\\newline
\textbf{Hybrid Approach} \\
The third approach is a mixture of the top-down and bottom-up approaches. The dataset annotators adapted an interactive approach to leverage high-level ideation from experts and the generation power of LLMs. In this process, the annotators start the LLM dialogue with task context, motivation, end goals and the codebase. In an iterative manner, the LLM is then prompted to elaborate and adapt its plan as the annotators increase complexity of the event history. Following this, the annotators have finer control over the areas in which complexity needs to be added which allows them to tune questions and situations accordingly, thereby making the process more flexible. Once the complete event history is generated, the annotator then manually attempts to solve the task to ensure completeness of the procedure.   

\paragraph{Dataset Statistics}
We provide a set of seven metrics to validate the quality of generated dataset, to ensure that the dataset is indeed diverse and includes long-term memory across multiple platforms, as designed. Detailed metric definitions and statistics are provided in~\autoref{tab:dataset_metrics}. % and~\autoref{fig:distribution_plots}.

\section{Experimental Setup}

\subsection{Methods} 

\textbf{\textsc{LLM+NoMem}}: This method has the LLM driving the agent without any within-agent memory components included.

\textbf{\textsc{LLM+Mem0}}: In this method, the agent consists of a LLM with an additional within-agent \textsc{Mem0}-based \cite{chhikara2025mem0}  memory component that exposes an additional set of its own tools, for example,  \textit{search\_memory()}, \textit{store\_memory()}, \textit{get\_memories()},  to the agent. We use \textsc{gpt-4o-mini}~\cite{hurst2024gpt} LLM-based embeddings and ChromaDB for the vector store in \textsc{Mem0}.

\textbf{\textsc{LLM+Zep}}: This has a similar agent architecture to the \textsc{LLM+Mem0} approach except that the memory component used is \textsc{Zep} \cite{rasmussen2025zeptemporalknowledgegraph}. We choose \textsc{Zep} as the second memory component since, having additional knowledge-graph based aspects to its internal memory component's architecture. 

For our memory experiments involving \textsc{Mem0} and \textsc{Zep}, we use the default settings from their respective APIs.

\subsection{Evaluation Measures} 

\textbf{Correctness:} For each of the questions $Q_i = (q_{i,1},q_{i,2} \ldots q_{i,n})$, our benchmark provides a set of respective expected answers ${A}_i^* = (a_{i,1}^*, a_{i,2}^* \ldots a_{i,n}^*)$. When the agent is given the tool-accessible timeline, it returns respective agent answers as it is prompted with each $q_i$ in order, yielding through its responses the agent answers ${A}_i = (a_{i,1}, a_{i,2} \ldots a_{i,n})$. Using a LLM-judge $Judge(q,a,a*)$ (\textsc{gpt-4o}), we output a Correctness score for each respective triple of question, expected answer and agent answer. The average of this score across all indices gives us the Correctness for an instance.

% \textbf{Efficiency:} The number of tool calls made to access elements of the timeline,  noted $TC$, is undesirable if it is excessive. To check for efficiency of tool calling, we measure it as $Eff = \frac{1}{e^\frac{|TC|-TC_{min}}{TC_{min}}}$ for $TC>=TC_{min}$ and $1$ for $TC<TC_{min}$.

\textbf{Efficiency:} The number of tool calls made to access elements of the timeline,  noted $TC$, is undesirable if it is excessive. To check for efficiency of tool calling, we measure it as $\text{Efficiency} = e^{-\frac{|TC|-TC_{min}}{TC_{min}}}$ for $TC>=TC_{min}$ and $1$ for $TC<TC_{min}$, where $TC_{min} = 10$.

\textbf{Redundancy:} Consider $TC_{i} = \{ {tcname_{i,k}}({tcargs_{i,k}})\}$ be the set of tool calls where $tcname_{i,k}$ and ${tcargs_{i,k}}$ are tool name and arguments respectively. We want to capture to what extent the agent is making tool calls (in the complete sense, including args) that are equivalent to or subsumed by tool calls made already i.e. ${TC}_{i,j} \subseteq {TC}_{i,k}$ s.t $j>k$ and are hence redundant. Measuring this is particularly important given we ask several questions. In this setting, maximally reusing the information obtained already while answering earlier questions is a desirable behavior.
    
Since this needs both soft matching  as well as noticing subsumption (e.g. \textit{get\_ticket(id="abc")} vs \textit{list\_all\_tickets())}, we prompt a LLM judge with our definition and the completely described TC, asking it to enumerate if each tool call is redudant given the previous and return the aggregate. Redundancy is defined as the fraction of $TC$ that is redundant.

Note that Efficiency and Redundancy are intrinsic metrics that measure the efficiency of the memory acquisition mechanism the model exhibits while doing the task, and higher Efficiency or lower Redundancy at the cost of a non-trivial drop in Correctness is in general not desirable.

\section{Results and Discussions}
A compilation of our results from our approaches with the frontier LLMs at the time of writing, namely \textsc{gpt-5} \cite{openaigpt5syscard} and \textsc{gemini-2.5-pro} \cite{comanici2025gemini} can be found in Table \ref{tab:memtrack_performance}.

We see that the \textsc{gpt-5} family based methods considerably outperforms the \textsc{gemini} family ones irrespective of the accompanying memory method used, reaching roughly four times the performance in terms of correctness.

%\subsection{How well do LLMs generally do, covered implicitly, answers Question 1}

\begin{table*}[!ht]
\caption{Results on {\datasetname} across methods recording the Correctness (averaged per-example over questions), Efficiency, Redundancy as well as extent and cross-platform entropy of tool calling. All  results are averaged across 5 runs and statistical significance is reported in~\autoref{tab:appendix_memtrack_correctness_sigma}.}
    \centering
    \small
    \setlength{\tabcolsep}{2pt}
    \begin{tabular}{l c c c c c}
        \toprule
         Method & Correct. ($\uparrow$)  & Eff. ($\uparrow$)  & Red. ($\downarrow$)  & $TC_{i} (mean/max)$ ($\uparrow$) & ($H(TC_{i})$) ($\uparrow$)   \\\midrule
         \textsc{gpt-5+NoMem}  & 0.601  & \textbf{0.667} & \textbf{0.206}  & 13.22/45.4 & 0.978   \\
         \textsc{gpt-5+Mem0}  & \textbf{0.610}  & 0.656 & 0.214  & 13.29/45.0 & \textbf{1.023}    \\
         \textsc{gpt-5+Zep}  & 0.601 & 0.660 & 0.214  & 13.23/\textbf{51.4} & 0.949  \\
         \textsc{gemini-2.5-pro+NoMem} & 0.144 & 0.638 & 0.237 & \textbf{13.72}/48.4 & 0.840   \\
         \textsc{gemini-2.5-pro+Mem0} & 0.118 & 0.658 & 0.240 & 12.53/46.8 & 0.838  \\
         \textsc{gemini-2.5-pro+Zep} & 0.140 & 0.662 & 0.238 & 12.51/47.8 & 0.747 \\
         %\textsc{claude-4.1-sonnet+NoMem} &  & xx & xx & xx  \\
         %\textsc{claude-4.1-sonnet+Mem0} & xx & xx & xx & xx  \\
         %\textsc{claude-4.1-sonnet+Zep} & xx & xx & xx & xx  \\
         %\textsc{Anthropic Claude-3.7-Sonnet\textsuperscript{*}} & xx & xx & xx & xx  \\
         %\textsc{Gemini-2.5-Pro-Preview-05-06}\textsuperscript{*\textdagger} & \textbf{xx} & \textbf{xx} & \textbf{xx} & xx  \\
         %\textsc{Gemini-2.5-Flash-Preview-04-17}\textsuperscript{*\textdagger} & xx & xx & xx & xx \\
         \bottomrule
    \end{tabular}
    \label{tab:memtrack_performance}
    %Insufficient context length is marked as \texttt{CLE}.
    %Models marked with \textsuperscript{*} have reasoning set to "high"; \textdagger\ indicates 1M+ token context window.
    %Pearson correlation b/w overall human and generated scores is shown under $\rho$.\protect\footnotemark
\end{table*}

\subsection{Quantitative Analysis}
\textbf{Do Agent Responses Invoke Tool Calls Successfully?}  All approaches show  high mean as well as median fraction of successful tool calls (tool calls which execute and return back) $\approx \in (0.90,0.96)$, with a minimum between $0.7$ and $0.8$ (exact numbers in Appendix Table \ref{tab:appendix_tool_calling_success}). On an average, \textsc{gemini-2.5-pro} . Though there is minor room for improvement here, both \textsc{gpt-5} and \textsc{gemini-2.5-pro} seem fairly fluent at this aspect. \\

\textbf{Do The Memory Components Improve Performance?} We see that both \textsc{Mem0} and \textsc{Zep} components do not cause any significant improvement in the performance of \textsc{GPT-5} as well as \textsc{Gemini-2.5-Pro}. Furthermore, in the case of \textsc{gemini-2.5-pro+Mem0}, we also notice a slight degradation in performance due to the introduction of \textsc{Mem0}.

\textbf{Does Performance Decay with Follow-Ups?:} We observe a small but notable drop in mean correctness when only follow-up questions are considered. This pattern holds consistently across all methods.

% \begin{table*}[!ht]
%     \centering
%     \small
%     \setlength{\tabcolsep}{4pt}
% \begin{tabular}{lcccccc} \hline 
% & \textbf{gpt-5+nomem} & \textbf{gemini+nomem} & \textbf{gpt-5+mem0} & \textbf{gemini+mem0} & \textbf{gpt-5+zep} & \textbf{gemini+zep} \\ \hline 
% \text{Overall} & 0.601 & 0.144 & 0.588 & 0.118 & 0.604 & 0.140 \\ 
% \text{Follow-Up} & 0.571 & 0.121 & 0.553 & 0.094 & 0.585 & 0.113 \\ \hline 
% \end{tabular} 
% \caption{Drop in performance on follow-up questions}
% \label{tab:followupdrop_perf}
% \end{table*}

\begin{table*}[!ht]
\caption{Drop in performance on follow-up questions}
    \centering
    \small
    \setlength{\tabcolsep}{4pt}
\begin{tabular}{lcccccc} \hline 
& \multicolumn{2}{c}{\textbf{\textsc{nomem}}} & \multicolumn{2}{c}{\textbf{\textsc{mem0}}} & \multicolumn{2}{c}{\textbf{\textsc{zep}}} \\ 
& \textbf{\textsc{gpt-5}} & \textbf{\textsc{gemini}} & \textbf{\textsc{gpt-5}} & \textbf{\textsc{gemini}} & \textbf{\textsc{gpt-5}} & \textbf{\textsc{gemini}} \\ \hline 
\text{Overall} & 0.601 & 0.144 & 0.588 & 0.118 & 0.604 & 0.140 \\ 
\text{Follow-Up} & 0.571 & 0.121 & 0.553 & 0.094 & 0.585 & 0.113 \\ \hline 
\end{tabular}
\label{tab:followupdrop_perf}
\end{table*}

%\subsection{Does Better Reasoning Help?}

%\subsection{Does post-training on our environment improve both self and generalized benchmark performance?}

\subsection{Qualitative analysis}

\subsubsection{Illustrative \textsc{GPT-5} Run}
In this example run, (of \textsc{GPT-5} solving the \textit{nemo\_run\_cli\_crisis} example), shown in full in Appendix \S\ref{subsec:appendix_illustrativeGPT5Run} , we observe how solving an instance in our environment makes a \textsc{GPT-5} driven agent richly navigate through making a total of 42 platform-interleaving tool calls spread over filesystem (magenta) , Linear (olive) and Slack (blue) before finally leading it to answer the final two follow-up questions [which are relatively phrased] correctly. At the outset, it also has to clone the repository of interest in its own filesystem, to facilitate the filesystem exploration and navigation. During this run, it has to switch between tool-calling against different platforms 7 times. (we show an excerpt here)

% \textcolor{olive}{list\_tickets()} $\rightarrow$ \textcolor{olive}{get\_tickets(ticket\_id = e8da2eed-bb7c-4d9f-93b2-89977fe48f2d)} \ldots (Getting More Tickets) $\rightarrow$ \textcolor{magenta}{search\_file\_content(pattern="", case\_sensitive = False, max\_matches=200, context\_lines=1)} $\rightarrow$ \textcolor{blue}{get\_unread\_messages()} $\rightarrow$ list\_channels() $\rightarrow$ \textcolor{blue}{get\_channel\_messages(channel="engineering", limit=100)} $\rightarrow$ \textcolor{olive}{get\_ticket(ticket\_id="171754fc-bca8-4057-8c5f-410ed9721590"} $\rightarrow$ Answer("2025-05-15") $\checkmark$ $\rightarrow$ Question 3 Asked $\rightarrow$ \textcolor{magenta}{list\_directory(path="NeMo-Run/nemo\_run", detailed=true)} $\rightarrow$ \textcolor{magenta}{search\_file\_content(pattern="32", search\_path= "NeMo-Run/nemo\_run")}  $\rightarrow$ Answer(142) $\checkmark$

\noindent\fbox{\begin{minipage}{\textwidth}
\small
\textcolor{olive}{list\_tickets()} $\rightarrow$ \textcolor{olive}{get\_tickets(ticket\_id = e8da2eed-bb7c-4d9f-93b2-89977fe48f2d)} $\ldots$ (Getting More Tickets) $\rightarrow$ \textcolor{magenta}{search\_file\_content(pattern="", case\_sensitive = False, max\_matches=200, context\_lines=1)} $\rightarrow$ \textcolor{blue}{get\_unread\_messages()} $\rightarrow$ list\_channels() $\rightarrow$ \textcolor{blue}{get\_channel\_messages(channel="engineering", limit=100)} $\rightarrow$ \textcolor{olive}{get\_ticket(ticket\_id="171754fc-bca8-4057-8c5f-410ed9721590")} $\rightarrow$ Answer("2025-05-15") $\checkmark$ $\rightarrow$ Question 3 Asked $\rightarrow$ \textcolor{magenta}{list\_directory(path="NeMo-Run/nemo\_run", detailed=true)} $\rightarrow$ \textcolor{magenta}{search\_file\_content(pattern="32", search\_path= "NeMo-Run/nemo\_run")} $\rightarrow$ Answer(142) $\checkmark$
\end{minipage}}

\subsubsection{Redundancy Patterns}
Given 20\% redundancy, we examine the sequence of tool calls made by models across examples as well as the reasoning outputs of our Redundancy LLM judge to ferret out some notable patterns. \\
\textbf{General-To-Specific Info Access Redundancy:} Agents often repeat up a general platform enlisting call (e.g. \textit{list\_tickets()}) by a more specific repeated  call (e.g. \textit{list\_tickets(limit=100)}).

% REDUNDANCY ANALYSIS EXAMPLE START
% memory_benchmark_zep_gpt5_results_20250829_233315
% config_langflow_composio_crisis
% run_2
% {'agent_computed_metrics': {'tool_calls_total': 32, 'tool_calls_failed': 4, 'tool_call_efficiency_rate': 0.875, 'idle_time_total_seconds': 140.48708510398865, 'max_unread_messages': 1, 'slack_exchanges': 7}, 'conversation_tool_calls': 32, 'efficiency_evaluation': {'score': 0.028115659748972045, 'is_correct': False, 'reasoning': 'Inefficient: 32 tool calls (> 7), score = 1/e^(32/7.0) = 0.028', 'evaluation_time': 3.1948089599609375e-05}, 'redundancy_evaluation': {'score': 0.84375, 'is_correct': True, 'reasoning': 'Redundancy: 5/32 calls redundant. The calls get_notifications({}) at indices 6, 24, and 25 are redundant as they retrieve the same information. The call list_tickets({}) at index 7 is redundant with list_tickets({"limit":100}) at index 28, as the latter is a broader query. The call notification_history({"limit":50,"offset":0}) at index 17 is redundant with notification_history({"limit":100,"offset":0}) at index 29, as the latter is a broader query.', 'evaluation_time': 2.037529945373535}}
% REDUNDANCY ANALYSIS EXAMPLE END

\textbf{Repeated Info Access After Interlude:} Given a sufficient gap of 3 or more turns, we notice a tendency to repeat information accessing tool calls E.g. a \textsc{gpt-5} based agent on the \textit{config\_plexe\_autocroll\_crisis} repeats \textit{get\_unread\_messages()} to list all slack messages as its 11th and 23rd tool calls and, in another run, \textit{get\_ticket(ticket\_id=x)} to re-access the same Linear ticket with id $x$ on its 8th and 25th tool call. This is not limited to Slack and Linear, in \textit{config\_vg\_12}, the file \textit{evals/elsuite/hr\_ml\_agent\_bench/benchmarks/cifar10} is read two times at the 9th and 17th tool calls made by the \textsc{gpt5+zep} agent. Some instances repeat file reads with overlapping line ranges.

% REDUNDANCY ANALYSIS EXAMPLE START
% memory_benchmark_zep_gpt5_results_20250829_233315
% config_browser_use_timeout_crisis
% run_2
% {'agent_computed_metrics': {'tool_calls_total': 59, 'tool_calls_failed': 19, 'tool_call_efficiency_rate': 0.6779661016949152, 'idle_time_total_seconds': 111.57132482528687, 'max_unread_messages': 1, 'slack_exchanges': 3}, 'conversation_tool_calls': 59, 'efficiency_evaluation': {'score': 0.0005940355282204004, 'is_correct': False, 'reasoning': 'Inefficient: 59 tool calls (> 7), score = 1/e^(59/7.0) = 0.001', 'evaluation_time': 3.314018249511719e-05}, 'redundancy_evaluation': {'score': 0.864406779661017, 'is_correct': True, 'reasoning': "Redundancy: 8/59 calls redundant. The redundant calls include multiple 'get_notifications({})' calls (3 times), repeated 'get_ticket' calls for the same ticket ID '3527ffee-103d-4a0d-85db-fe574b941170' (2 times), and repeated 'read_file' calls with overlapping line ranges for '/home/repo/browser_use/config.py' (3 times).", 'evaluation_time': 2.67514705657959}}
% REDUNDANCY ANALYSIS EXAMPLE END

\textbf{Progressively Widening Exploration:} In some cases, the agent chooses to explore a platform very gradually with increasing limits, if the tool parameters allow this e.g. \textit{get\_channel\_messages(channel='backend',limit=50)} followed by limit 100. Agents being conservative in this regard is justified, so this is not an anti-pattern unless its overdone to  point of small increments.

\subsection{Answering Research Questions}

\textbf{RQ1: Can modern LLMs reason over large event histories and codebases to answer SWE questions?} As seen in~\autoref{tab:memtrack_performance}, frontier LLM families such as \textsc{gpt-5} exhibit suboptimal performance on the \datasetname ($\approx 60\%$) irrespective of memory components. This suggests that there is a significant room for improvement, particularly on improving large context reasoning and understanding of follow-up questions as seen in Table ~\autoref{tab:followupdrop_perf}. SoTA models are unable to properly utilize memory implementations and addition of memory tools leads to a clear increase in redundancy of tool calls. Finally, the increase in tool call entropy as displayed in~\autoref{tab:memtrack_performance} suggests that diverse information gathering using tools helps answer questions more reliably.

% The follow-up correctness being lower, as seen from ~\autoref{tab:followupdrop_perf} indicates that there is even further room to improve on how well 
\textbf{RQ2: Can agents use memory databases and backends such as \textsc{Zep} and \textsc{Mem0} effectively to reason over large codebases?} According to \autoref{tab:memtrack_performance}, we can observe that memory equipped LLMs fail to call memory tools effectively. LLMs with memory tools consistently display increased redundancy as well as drop in performance efficiency. As observed in the qualitative analysis above, models generally prefer repeatedly accessing information over using the memory component.

\textbf{RQ3: Can LLM agents optimally access and remember information from across platforms?}  
We observe that retention of cross-platform information is suboptimal and models need to access information repeatedly. Given a $\ge20\%$ redundancy in tool calls provides support to this suboptimality. Additionally, as observed in~\autoref{tab:followupdrop_perf}, LLMs display poor follow-up performances suggesting their inability to remember and utilize cross platform information effectively. Finally, the tool call entropy displayed in ~\autoref{tab:memtrack_performance} suggests that diverse tool use is still a limiting factor improving on which can drive up performance for current approaches. Thus, the lack of saturation on this metric shows the gap between memory assimilation across multiple threads.

\section{Conclusion \& Future Work}
By combining manual, bottom-up, and hybrid data curation approaches, we developed {\datasetname}, a benchmark to test agent memory in a multi-platform  environments that  faithfully reflect real organizational workflows. Our experiments show that while frontier LLMs like \textsc{GPT-5} exhibit some capacity to reason across large event histories, performance declines on follow-up queries. Memory components such as \textsc{Mem0} and \textsc{Zep} provide limited gains, and tool-use to access memory remains inefficient. Our framework provides an umbrella to build future environments where the agent is not only given access to a rich set of events, but furthermore  contextually act, such as creating Linear tickets, sending Slack messages, to get involved in the future unfolding of the organization timeline and aim to drive the overall organization to complete the tasks. We avoided exploring this harder class of settings in this initiating work to focus on crafting an ecologically valid base. In the future, \datasetname  can be extended to other domains such as marketing or sales that commonly involve a large overlap of internal and external communication contexts. 

\bibliography{custom}
\bibliographystyle{unsrtnat}  % numeric, ordered by citation

\appendix

\section{Prompts}
\label{sec:appendix_prompts}

\subsection{Dataset Generation}

\subsubsection{Bottom-Up Approach}
\label{appendix:bottom-up}
\begin{lstlisting}[
  breaklines=true,
  frame=single,
  basicstyle=\small\ttfamily,
  columns=flexible
]
 Refer to the task details attached here and create a situationally aware dataset that tests LLM's memory skills. Be comprehensive, careful (verify your steps) and optimal while creating the dataset. Analyze the given examples, especially their questions and expected answers to understand the level of complexity and cross platform communication that is expected. You must run the script and reiterate the process in case the models pass all tests (i.e. the dataset is too easy for the model). You must always verify that the model fails on your dataset and you must carefully iterate while making sure that you don't lose track of the task and trajectory that you want to follow to get to the answer. Always be ecologically valid when creating questions and think from the perspective of actual user requirements and complexity.   

 {task_details}
\end{lstlisting}

The task description is as follows:
\begin{lstlisting}[
  breaklines=true,
  frame=single,
  basicstyle=\small\ttfamily,
  columns=flexible
]
Configuration Setup
Before beginning, configure the following variables:

OWNER: The GitHub repository owner (e.g., "openai", "microsoft", "google")
REPO: The repository name (e.g., "mle-bench", "vscode", "tensorflow")
ORGANIZATION: The organization name for internal communications (can be same as OWNER or different)

Creating the Setup
Step 1: Repository Analysis
Analyze the target repository: https://github.com/{OWNER}/{REPO}.git
Pull all solved issues using the following script:
bashCopycurl -H "Accept: application/vnd.github.v3+json" \
     -H "User-Agent: curl" \
     "https://api.github.com/repos/{OWNER}/{REPO}/issues?state=closed" | jq -r '.[] | .number, .title, ""'
Step 2: Issue Investigation
Identify interesting issues that involve technical discussions, feature implementations, or bug resolutions. For promising issues, pull detailed information:
bashCopycurl -H "Accept: application/vnd.github.v3+json" \
     -H "User-Agent: curl" \
     "https://api.github.com/repos/{OWNER}/{REPO}/issues/{ISSUE_NUMBER}" | jq -r '.title, "", .body'
Pull the complete comment history:
bashCopycurl -H "Accept: application/vnd.github.v3+json" \
     -H "User-Agent: curl" \
     "https://api.github.com/repos/{OWNER}/{REPO}/issues/{ISSUE_NUMBER}/comments" | \
jq -r '.[] | "Author: " + .user.login + "\nDate: " + .created_at + "\n" + .body + "\n" + ("-" * 50) + "\n"'
Creating the Multi-Platform Scenario
Core Requirements for Task Difficulty
The scenario must be designed such that:

No single platform contains the complete answer - Information must be distributed across Slack, Linear, and potentially email/documents
Minimum 5 reasoning hops required - Each hop should require connecting information from different sources
Minimum 3 platform switches - Must require switching between Slack channels, Linear tickets, and other platforms
Temporal dependencies - Some information should only be meaningful when combined with timestamps from different platforms

Multi-Platform Information Distribution Strategy
Structure your scenario with information scattered as follows:
Platform 1 (Slack #engineering): Contains initial problem identification and high-level technical approach decisions
Platform 2 (Linear tickets): Contains detailed implementation specifications, status updates, and technical requirements
Platform 3 (Slack #random or #general): Contains casual mentions of blockers, dependencies, or context that affects the technical decision
Platform 4 (Slack DMs or #leadership): Contains resource allocation decisions, priority changes, or approval processes
Platform 5 (Additional tickets/email): Contains external dependencies, vendor communications, or compliance requirements
Scenario Creation Guidelines

Create realistic organizational personas with distinct roles:

Engineering leads, senior developers, product managers
DevOps, security, QA team members
External stakeholders or vendor contacts


Implement strategic information fragmentation:

Key decision rationale split across multiple Slack threads
Implementation details spread between Linear descriptions and comments
Critical context buried in seemingly unrelated conversations
Timeline information requiring cross-referencing multiple platforms


Add high-quality distractors:

Similar but unrelated technical discussions
Multiple tickets with similar names or purposes
Conversations about abandoned approaches or rejected solutions
Time-overlapping discussions about different features


Ensure temporal complexity:

Decisions that get revised across platforms
Information that becomes relevant only after later context
Status changes that affect interpretation of earlier discussions



Question Design Requirements
Ecological Validity Principles
Questions should reflect real workplace information needs:
Instead of: "What was the exact configuration parameter set for the Redis cache timeout?"
Design: "I'm trying to understand why our caching layer behaves the way it does. Can you help me figure out what drove the performance decisions we made last quarter?"
Instead of: "Which team member proposed the authentication fix?"
Design: "The client is asking about our security implementation timeline. Who should I credit for the breakthrough that got us past that login issue?"
Instead of: "What was the final status of ticket LIN-403?"
Design: "I need to prep for the retrospective meeting. Can you remind me how that problematic feature request got resolved?"
Strategic Vagueness Implementation
Questions should be contextually ambiguous to require genuine reasoning:
Time-based Vagueness

"That issue from a few weeks back" (multiple issues in timeframe)
"The recent performance discussion" (several performance topics)
"After the client escalation" (need to identify which escalation)

Domain-based Vagueness

"The API changes" (multiple API modifications)
"The security fix" (several security-related tickets)
"The database optimization" (multiple DB-related work streams)

Role-based Vagueness

"What the team decided" (need to identify which team/decision)
"The solution Alex proposed" (Alex proposed multiple solutions)
"The approach we ultimately went with" (multiple approaches discussed)

Outcome-based Vagueness

"The workaround that actually worked" (multiple workarounds attempted)
"The change that fixed the production issue" (multiple changes made)
"The configuration that resolved the bottleneck" (several configs modified)

Multi-Hop Reasoning Structure
Your question must require connecting information such as:
Hop 1: Vague question reference -> Identify specific context period/domain
Hop 2: Context clues -> Locate initial problem discussion (Slack #engineering)
Hop 3: Problem details -> Find corresponding implementation ticket (Linear)
Hop 4: Implementation ticket -> Discover blocking dependency (Slack #random)
Hop 5: Dependency context -> Locate resolution approach (Different Linear ticket)
Hop 6: Resolution approach -> Confirm final outcome (Slack #leadership)
Example Ecologically Valid + Vague Questions
Scenario: Performance optimization issue resolution
Poor Question: "What caching strategy did ticket LIN-427 implement on March 15th?"
Good Question: "Hey, I'm onboarding the new performance engineer and they're asking about that caching situation we dealt with after the big slowdown. Can you help me explain what we ended up implementing and why it worked?"
Required reasoning path:

Identify "big slowdown" -> find incident discussion in #engineering
Locate related performance tickets in Linear
Find caching strategy discussions across multiple threads
Connect implementation approach to success metrics
Verify final solution through status updates and retrospective comments

Scenario: Security vulnerability fix
Poor Question: "Who fixed the authentication bug in issue #234?"
Good Question: "The security team is doing their quarterly review and asked about that authentication problem that had everyone stressed. Can you remind me what the root cause turned out to be and how we prevented it from happening again?"
Required reasoning path:

Identify "authentication problem" from stress context clues
Find initial security discussions in relevant channels
Locate corresponding security tickets and investigations
Connect root cause analysis across multiple conversations
Identify prevention measures from implementation updates

Scenario: Feature rollback decision
Poor Question: "Why was the notification feature disabled?"
Good Question: "I'm updating the roadmap document and need to explain the notification feature situation to stakeholders. What ended up being the main concern that led to our decision there?"
Required reasoning path:

Identify notification feature context from roadmap implications
Find feature development discussions
Locate rollback decision conversations
Connect stakeholder concerns across multiple platforms
Verify decision rationale through leadership communications

Progressive Question Design
Primary Question (Multi-hop reasoning)
Start with an ecologically valid, appropriately vague question requiring the full reasoning chain.
Follow-up Questions (Memory testing)
Design 2-3 progressively specific follow-ups that test retention:
Follow-up 1: "And what was the timeline concern that influenced that decision?"
Follow-up 2: "Who were the key people who made that call?"
Follow-up 3: "How did this connect to the client requirements we discussed?"
Output Requirements
JSON Structure
Create a comprehensive event history JSON containing:

Chronologically ordered events across all platforms
Realistic message threading and references
Proper user attribution and timestamps
Status transitions for Linear tickets
Cross-platform references and mentions

Question-Answer Pairs
Design questions that:

Sound like actual workplace inquiries - colleagues asking for context, preparing for meetings, onboarding explanations
Require domain knowledge to interpret - understanding what "the performance issue" or "the client escalation" refers to
Have deterministic answers despite vague framing
Test both reasoning and memory through progressive follow-ups

Verification Criteria

Questions reflect realistic workplace information needs
Vagueness requires genuine reasoning rather than keyword matching
Answer requires information synthesis from minimum 3 platforms
No single message thread or ticket contains sufficient information
Human evaluator with JSON access can definitively verify correctness

File Deliverables

event_history_{REPO}.json - Complete multi-platform scenario
config_{REPO}.yaml - Question-answer configuration with ecological validity ratings
verification_guide.md - Step-by-step reasoning path for each question

Remember: The goal is to create questions that sound like natural workplace communication while requiring sophisticated multi-platform reasoning to answer accurately.
\end{lstlisting}

The outputs are then iteratively made more difficult if the evaluated agent passes the questions around the event history. Difficulty can be increased along the dimension of either question complexity or by making the event history more complicated. We explore both cases on a per-sample basis to ensure validity of the dataset.

\subsubsection{Top-Down Approach}

The system prompt is as below:
\begin{lstlisting}[
  breaklines=true,
  frame=single,
  basicstyle=\small\ttfamily,
  columns=flexible
]
""" you are a data generator to create a timeline in the format as this example: 

Example timeline:{timeline_examples}.

Your task is as below:

{task_description}

Only output the timeline in JSON format.
"""
\end{lstlisting}

This is an example of task description in natural language:  
\begin{lstlisting}[
  breaklines=true,
  frame=single,
  basicstyle=\small\ttfamily,
  columns=flexible
]
Create a timeline, starting from the PM Alice who wants to build a product feature that allows searching product images based on descriptions. The engineer bob created a linear ticket to list the step of implementation. Alice and bob started a slack message and confirmed that the search is based on the keyword of the product description. This is further confirmed with Diana in another thread between Diana and bob. The initial implementation was done with a few git commits and the original ticket was closed. But when Alice tested the product feature, she reported that a lot of times, the product cannot be found based on here description. Bob, Diana and Alice started another Slack thread, and after 20 messages, they found that what Alice meant by description is from users, not directly from product description in their cataglog. They decided to implement the whole system differently, and create a new linear ticket for it.
\end{lstlisting}

\subsection{Prompts and Interesting  for Evaluation Measures}
\label{subsec:appendix_evalPrompts}

\subsubsection{Correctness}
\label{subsubsec:appendix_correctness}

\begin{lstlisting}[
  breaklines=true,
  frame=single,
  basicstyle=\small\ttfamily,
  columns=flexible
]
"""You are evaluating whether an AI agent correctly answered a memory-related question about a
workplace timeline.

QUESTION: {question}

EXPECTED ANSWER: {expected_answer}

AGENT'S EXTRACTED ANSWER: {extracted_answer}

Your task is to determine if the agent's answer is correct. Consider:
1. Exact matches are obviously correct
2. Semantic equivalence (same meaning, different wording)
3. Partial correctness (got part of a multi-part answer right)
4. Reasonable interpretations of ambiguous questions

Respond with ONLY a JSON object in this exact format:
{{
    "score": <float between 0.0 and 1.0>,
    "is_correct": <true or false>,
    "confidence": <float between 0.0 and 1.0>,
    "reasoning": "<brief explanation of your evaluation>"
}}

Examples:
- If expected "done" and agent said "completed": score 1.0, correct true
- If expected "alice, bob" and agent said "alice": score 0.5, correct false
- If expected "in_progress, charlie" and agent said "in_progress, urgent, charlie": score 1.0, correct true
  (extra info OK)
- If completely wrong: score 0.0, correct false"""
\end{lstlisting}

\subsubsection{Redundancy}
\label{subsubsec:appendix_efficiency}

\begin{lstlisting}[
  breaklines=true,
  frame=single,
  basicstyle=\small\ttfamily,
  columns=flexible
]
"""Analyze these tool calls for memory retrieval redundancy:

{chr(10).join(formatted_calls)}

Task: Identify redundant tool calls where the agent re-retrieves similar information.

Consider these as redundant:
- Same tool with very similar parameters (e.g., "get_ticket(id1)" then "get_ticket(id1)")
- Semantically equivalent queries (e.g., "list_tickets(status='done')" then "list_tickets(status='completed')")
- Broad queries that contain info from specific queries (e.g., "get_ticket(id1)" then "list_tickets()")

Consider these as NOT redundant:
- Different tools entirely
- Same tool with meaningfully different parameters
- Natural progression from general to specific

Respond with ONLY this JSON format:
{{
    "redundant_calls": <number of redundant calls>,
    "total_calls": <total number of calls>,
    "redundancy_fraction": <redundant_calls / total_calls>,
    "efficiency_score": <1.0 - redundancy_fraction>,
    "reasoning": "<brief explanation of which calls were redundant>"
}}
"""
\end{lstlisting}

\subsection{Prompts for Statistics}
\label{subsec:appendix_statistics}

\subsubsection{Cross Reference}
\label{subsubsec:appendix_crossreference}

\begin{lstlisting}[
  breaklines=true,
  frame=single,
  basicstyle=\small\ttfamily,
  columns=flexible
]
"""
Analyze this {current_platform} content for references to other platforms.
Rate each reference type from 0.0 to 1.0:

Content: "{content}"

Rate how much this content references or alludes to:
- Linear tickets/issues (mentions like "ticket", "Linear", "BACK-456", "milestone", "bug report")
- Slack messages/channels (mentions like "channel", "message", "thread", "@someone said", "in #engineering")
- Git commits/PRs/code (mentions like "commit", "PR", "pull request", "branch", "merge", "code change")

Respond only with a JSON object:
{{"linear": 0.0, "slack": 0.0, "git": 0.0}}

Examples:
- "Created BACK-456 for the TypedDict issue" -> {{"linear": 0.9, "slack": 0.0, "git": 0.0}}
- "PR #12124 is already getting review" -> {{"linear": 0.0, "slack": 0.0, "git": 0.8}}
- "@alex_chen mentioned in #engineering" -> {{"linear": 0.0, "slack": 0.8, "git": 0.0}}
"""
\end{lstlisting}

\section{Additional Results \& Analysis}

\subsection{Illustrative GPT-5 Run - Complete Example}
\label{subsec:appendix_illustrativeGPT5Run}

\textbf{Q1:} \textit{ Mike mentioned something about a default value he set in model configuration. Sarah referenced 'the specific case from the ticket' but never stated what it was. David mentioned finding 'the issue' but was vague about location details. By correlating across all these conversations and investigating the codebase, determine: what is the sum of all numeric values that appear as default parameters in factory function definitions within the test files?} $A^*$: 547 \\
\textbf{Q2:} \textit{The team communicated about a bug resolution timeline with vague date references like 'today', 'last month', and 'this afternoon'. By parsing all temporal references in context and correlating with the event timeline, determine: on what date did Sarah first report the factory function conversion issue?} $A^*$: 20250515 \\
\textbf{Q3:} \textit{Multiple conversations referenced configuration values, default parameters, and batch sizes across different contexts. By correlating \textbf{Mike's early comment about batch size} with the actual repository structure at the \textbf{specified commit}, determine: what is the exact line number where the first occurrence of the number 32 appears in any Python file within the nemo\_run directory?} $A^*$: 142 \\
\textbf{Tool Calls Made:} 

\noindent\fbox{\begin{minipage}{\textwidth}
\small
\textcolor{red}{list\_remote\_git\_repositories()} $\rightarrow$ \textcolor{brown}{run\_shell\_command(git clone http://git.local:3000/pgym/NeMo-Run.git)} $\rightarrow$ \textcolor{magenta}{list\_directory(path = NeMo-Run, detailed = true)} $\rightarrow$ \textcolor{magenta}{list\_directory(path = NeMo-Run/test, detailed = true)} $\rightarrow$ \textcolor{magenta}{read\_file(path = NeMo-Run/test/dummy\_factory.py)} $\rightarrow$ \textcolor{magenta}{search\_file\_content(pattern = \text{"@run.cli.factory"}, search\_path = "NeMo-Run/test", case\_sensitive = false, max\_matches = 100, context\_lines = 2)} $\rightarrow$ \textcolor{magenta} {read\_file(file\_path = NeMo-Run/test/cli/test\_api.py, start\_line=1, max\_lines = 1000)} $\rightarrow$  $\ldots$ $\ldots$ \textcolor{magenta}{(Further Navigation and Reading of Files)} $\rightarrow$ Answer(4858.9001) \ding{56}  $\rightarrow$   Question 2 Asked $\rightarrow$ \textcolor{olive}{list\_tickets()} $\rightarrow$ \textcolor{olive}{get\_tickets(ticket\_id = e8da2eed-bb7c-4d9f-93b2-89977fe48f2d)} $\rightarrow$ \textcolor{olive}{get\_tickets(ticket\_id="39b5f839-d79c-438c-8eb3-fee9bb5af641")} $\rightarrow$ \textcolor{olive}{get\_tickets(ticket\_id="970a53b2-ed24-4719-8ee5-153d03c3d983")} $\rightarrow$ \textcolor{magenta}{search\_file\_content(pattern="", case\_sensitive = False, max\_matches=200, context\_lines=1)} $\rightarrow$ \textcolor{blue}{get\_unread\_messages()} $\rightarrow$ list\_channels() $\rightarrow$ \textcolor{blue}{get\_channel\_messages(channel="engineering", limit=100)} $\rightarrow$ \textcolor{blue}{get\_ticket(ticket\_id="171754fc-bca8-4057-8c5f-410ed9721590")} $\rightarrow$ Answer("2025-05-15") $\checkmark$ $\rightarrow$ Question 3 Asked $\rightarrow$ \textcolor{magenta}{list\_directory(path="NeMo-Run/nemo\_run", detailed=true)} $\rightarrow$ \textcolor{magenta}{search\_file\_content(pattern="32", search\_path= "NeMo-Run/nemo\_run")}  $\rightarrow$ Answer(142) $\checkmark$
\end{minipage}}

% | Method | tool_call_success_rate(mean) | tool_call_success_rate(median) | tool_call_success_rate(min) |
% |---|---|---|---|
% | memory_benchmark_zep_gpt5_results_20250829_233315 | 0.9166253421574565 | 0.9698363343580734 | 0.7228571428571429 |
% | memory_benchmark_mem0_gemini_results_20250830_123405 | 0.9341399715496856 | 0.9535463659147869 | 0.7569696969696971 |
% | memory_benchmark_nomemory_gpt5_results_20250829_222210 | 0.9179321799852482 | 0.9730578155374758 | 0.6967320261437908 |
% | memory_benchmark_mem0_gpt5_results_20250829_233011 | 0.9306936590485259 | 1.0 | 0.7016932431509189 |
% | memory_benchmark_mem0_claude_results_20250829_233021 | 0.9523943646846241 | 0.9765930483464569 | 0.7717170821926236 |
% | memory_benchmark_zep_claude_results_20250829_233311 | 0.9106069620437194 | 0.9082030387912741 | 0.7026839826839827 |
% | memory_benchmark_nomemory_claude_results_20250829_232558 | 0.9531702448298721 | 0.9853937806808839 | 0.7761352804958679 |
% | memory_benchmark_nomemory_gemini_results_20250830_123311 | 0.9304976155395601 | 0.9491105430329569 | 0.71 |
% | memory_benchmark_zep_gemini_results_20250830_123111 | 0.9290553213977789 | 0.9534090909090909 | 0.645483040273736 |

\begin{table}[htbp]
\small
\centering
\caption{Tool Call Success Rates (SRs) Across Different Memory Benchmark Methods}
\begin{tabular}{lccc}
\toprule
\textbf{Method} & \textbf{Mean SR} & \textbf{Median SR} & \textbf{Min SR} \\
\midrule
zep\_gpt5\_results\_20250829\_233315      & 0.9166 & 0.9698 & 0.7229 \\
mem0\_gemini\_results\_20250830\_123405   & 0.9341 & 0.9535 & 0.7570 \\
nomemory\_gpt5\_results\_20250829\_222210 & 0.9179 & 0.9731 & 0.6967 \\
mem0\_gpt5\_results\_20250829\_233011     & 0.9307 & 1.0000 & 0.7017 \\
mem0\_claude\_results\_20250829\_233021   & 0.9524 & 0.9766 & 0.7717 \\
zep\_claude\_results\_20250829\_233311    & 0.9106 & 0.9082 & 0.7027 \\
nomemory\_claude\_results\_20250829\_232558 & 0.9532 & 0.9854 & 0.7761 \\
nomemory\_gemini\_results\_20250830\_123311 & 0.9305 & 0.9491 & 0.7100 \\
zep\_gemini\_results\_20250830\_123111   & 0.9291 & 0.9534 & 0.6455 \\
\bottomrule
\end{tabular}
\label{tab:appendix_tool_calling_success}
\end{table}

\subsection{Hyperparameters}
We use a temperature of 1 for all our experiments. We choose a higher temperature to allow reasoning and also to prevent an implicit length horizon from getting in the way of the relatively longer output token horizons and turn counts needed to solve the task.

\subsection{$\sigma$ of Correctness}

\begin{table*}[!ht]
\caption{$\sigma$ on Correctness numbers, averaged across 5 runs.}
    \centering
    \small
    \setlength{\tabcolsep}{2pt}
    \begin{tabular}{l c }
        \toprule
         Method & $\sigma$(Correctness)    \\\midrule
         \textsc{gpt-5+NoMem}  & 0.0594     \\
         \textsc{gpt-5+Mem0}  & 0.0610     \\
         \textsc{gpt-5+Zep}  & 0.0452   \\
         \textsc{gemini-2.5-pro+NoMem} & 0.0480    \\
         \textsc{gemini-2.5-pro+Mem0} & 0.0342  \\
         \textsc{gemini-2.5-pro+Zep} & 0.0430  \\
         %\textsc{claude-4.1-sonnet+NoMem} &  & xx & xx & xx  \\
         %\textsc{claude-4.1-sonnet+Mem0} & xx & xx & xx & xx  \\
         %\textsc{claude-4.1-sonnet+Zep} & xx & xx & xx & xx  \\
         %\textsc{Anthropic Claude-3.7-Sonnet\textsuperscript{*}} & xx & xx & xx & xx  \\
         %\textsc{Gemini-2.5-Pro-Preview-05-06}\textsuperscript{*\textdagger} & \textbf{xx} & \textbf{xx} & \textbf{xx} & xx  \\
         %\textsc{Gemini-2.5-Flash-Preview-04-17}\textsuperscript{*\textdagger} & xx & xx & xx & xx \\
         \bottomrule
    \end{tabular}
    \label{tab:appendix_memtrack_correctness_sigma}
    %Insufficient context length is marked as \texttt{CLE}.
    %Models marked with \textsuperscript{*} have reasoning set to "high"; \textdagger\ indicates 1M+ token context window.
    %Pearson correlation b/w overall human and generated scores is shown under $\rho$.\protect\footnotemark
\end{table*}

\subsection{Tool/Component Specifications}\label{subsec:appendix_toolspecifications}

\subsubsection{Slack}
\begin{enumerate}
    \item {\color{blue}$get\_unread\_messages(limit: int = 50) \rightarrow str$} - Retrieve unread messages grouped by channel/DM, automatically marked as read after retrieval
    \item {\color{blue}$send\_channel\_message(channel: str, message: str) \rightarrow str$} - Send message to a channel with @username mention support
    \item {\color{blue}$send\_direct\_message(to\_user: str, message: str) \rightarrow str$} - Send direct message to another user
    \item {\color{blue}$get\_channel\_messages(channel: str, limit: int = 50, after\_id: str = None) \rightarrow str$} - Retrieve channel messages with pagination support
    \item {\color{blue}$get\_direct\_messages(with\_user: str, limit: int = 50, after\_id: str = None) \rightarrow str$} - Retrieve DM conversation history with pagination
    \item {\color{blue}$list\_channels() \rightarrow str$} - List all available channels in workspace
    \item {\color{blue}$list\_users() \rightarrow str$} - List all users in the Slack workspace
    \item {\color{blue}$send\_offline\_message(to\_user: str, message: str) \rightarrow str$} - Send offline message when away from desk
\end{enumerate}

\subsubsection{Linear}
\begin{enumerate}
    \item {\color{olive}$create\_ticket(title: str, description: str, team: str, ...) \rightarrow str$} - Create new ticket with comprehensive metadata including priority, dates, labels, and milestones
    \item {\color{olive}$update\_ticket(ticket\_id: str, ...) \rightarrow str$} - Update existing ticket fields with authorization checks
    \item {\color{olive}$get\_ticket(ticket\_id: str) \rightarrow str$} - Retrieve comprehensive ticket details by ID
    \item {\color{olive}$list\_tickets(team: str = None, status: str = None, ...) \rightarrow str$} - Query tickets with filtering by team, status, lead, milestone, priority, and date ranges
    \item {\color{olive}$delete\_ticket(ticket\_id: str) \rightarrow str$} - Permanently remove ticket (admin-only operation)
    \item {\color{olive}$assign\_ticket(ticket\_id: str, lead: str) \rightarrow str$} - Assign or reassign ticket ownership with notifications
    \item {\color{olive}$add\_team\_member(team: str, member: str) \rightarrow str$} - Add user to ML or Engineering team (admin-only)
    \item {\color{olive}$remove\_team\_member(team: str, member: str) \rightarrow str$} - Remove user from team (admin-only)
    \item {\color{olive}$list\_team\_members(team: str = None) \rightarrow str$} - List team membership for specified or all teams
    \item {\color{olive}$create\_milestone(name: str, description: str, start\_date: str, end\_date: str) \rightarrow str$} - Create project milestone with unique naming (admin-only)
    \item {\color{olive}$update\_milestone\_progress(milestone: str, completion\_rate: float) \rightarrow str$} - Update milestone completion and sync to associated tickets (admin-only)
    \item {\color{olive}$list\_milestones() \rightarrow str$} - Retrieve all milestones with current progress status
\end{enumerate}

\subsubsection{Git}
\begin{enumerate}
    \item $list\_remote\_git\_repositories() \rightarrow str$ - List available remote repositories with clone instructions and commit information
\end{enumerate}

\subsubsection{Shell Commands}
The agent also has access to its own filesystems into which it can execute shell commands such as git clone as well as \textit{list\_directory()}, \textit{read\_file()}, \textit{search\_file()} etc

% \subsubsection{Docker Network}
% \begin{enumerate}
%     \item No tools provided - Docker network component handles container isolation infrastructure without exposing user-facing tools
% \end{enumerate}

\subsection{Memory Based Components}
\label{subsec:appendix_memorybasedcomponents}
\begin{enumerate}
    \item {\color{purple}$store\_memory(content: str, metadata: dict = None) \rightarrow str$} - Store important information in persistent memory for future recall and context building
    \item {\color{purple}$search\_memory(query: str, limit: int = 5) \rightarrow str$} - Search for contextually relevant memories using hybrid semantic and keyword search
    \item {\color{purple}$get\_memories(limit: int = 10) \rightarrow str$} - Retrieve recent stored memories for context awareness and debugging
\end{enumerate}

% \subsection{External/Platform Components}
% \label{subsec:appendix_externalplatformcomponents}

\definecolor{lightgray}{RGB}{245,245,245}

\newtcolorbox{apibox}[1]{
  colback=lightgray,
  colframe=black,
  fontupper=\scriptsize\ttfamily,  % Changed to scriptsize
  title=#1,
  arc=0pt,
  boxrule=0.5pt,
  before upper={\parindent=0pt},
}

\lstset{
  basicstyle=\scriptsize\ttfamily,  % Changed to scriptsize
  breaklines=true,
  postbreak=\mbox{\textcolor{red}{$\hookrightarrow$}\space},
}

\subsection{Compute Resources}
The following configuration was used for running evaluation experiments on \datasetname.

\begin{table}[h]
\caption{System Requirements to run environment}
\centering
\begin{tabular}{ll}
\toprule
\textbf{Resource} & \textbf{Requirement} \\
\midrule
Memory & 2GB per concurrent benchmark (Docker containers + memory providers) \\
Ports & Ensure 3000-3999 range is available \\
Docker Networks & System should support 10+ custom networks \\
\bottomrule
\end{tabular}
\label{tab:system_requirements}
\end{table}

\end{document}